\documentclass[prodmode,acmtecs]{acmsmall} 

\usepackage[ruled]{algorithm2e}
\usepackage{multicol}

\SetAlFnt{\small}
\SetAlCapFnt{\small}
\SetAlCapNameFnt{\small}
\SetAlCapHSkip{0pt}
\IncMargin{-\parindent}

\acmVolume{ }
\acmNumber{29 }
\acmArticle{ }
\acmYear{2015}
\acmMonth{12}

\doi{0000001.0000001}

\issn{1234-56789}

\begin{document}

\markboth{K. Eswaran and K.Damodhar Rao}{On non-iterative training of a neural classifier}

\title{On non-iterative training of a neural classifier}
\author{K. ESWARAN
\affil{Sreenidhi Institute of Science and Technology}
K. DAMODHAR RAO
\affil{Sreenidhi Institute of Science and Technology}
}
\begin{abstract}
Recently an algorithm, was discovered, which separates points in n-dimension
by planes in such a manner that no two points are left un-separated
by at least one plane{[}1-3{]}. By using this new algorithm we show
that there are two ways of classification by a neural network, for
a large dimension feature space, both of which are non-iterative and
deterministic. To demonstrate  the power of both these methods we apply them exhaustively to the classical pattern recognition problem: The Fisher-Anderson's, IRIS flower data set and present the results. 

It is expected these methods will now be widely used for the training of neural networks for Deep Learning not only because of their non-iterative and deterministic nature but also because of their efficiency and speed and will supersede other classification methods which are iterative in nature and rely on error minimization.
\end{abstract}

%
%

%
%



\acmformat{K.Eswaran and K.Damodhar Rao, 2015.On non-iterative training of a neural classifier.}

\begin{bottomstuff}

Author's addresses: K.Eswaran {and} K.Damodhar Rao, Department of Computer Science,
Sreenidhi Institute of Science and Technology,
Yamnampet, Ghatkesar, Hyderabad, 500010 India. 
\end{bottomstuff}

\maketitle

\begin{multicols}{2}
\section{Introduction}
Many researchers, from statisticians and neural network scientists
have long been trying to separate clusters by planes or discriminant
functions. This has been ever since the time of Fisher {[}4{]} and
Mahalanobis {[}5{]} and McCullough and Pitts{[}6{]}, Kolmogorov {[}7{]},
Werbos{[}8{]}, and Rumelhart, Hinton and Williams {[}9{]} and others
{[}10{]}, also the Deep Learning people{[}11{]}. But all have had
great difficulties in large n-dimensional space and it has been found
that it is not easy. So the phrases like ``The Curse
of Dimensionality'' and ``NP Hard Complexity'' 
have become a part of folk lore. However, from the very beginning
it was never very easy to separate clusters by planes. This is mostly
because, a cluster is NOT well defined, every cluster has its own
shape and in n-dimensions, one could have long thin filaments and
all kinds of snake like dragon like shapes, which constitute clusters.
Though statisticians try to approximate the shape of each cluster
as an ellipsoid or even as a simple sphere, clusters in general would
require more parameters to mathematically correctly define their shapes
than the number of coefficients required to define the planes which
are supposed to separate them. So all along it was, perhaps, very
naïve of all of us to have tried to separate clusters when such entities
are not mathematically well defined. It is far better to separate
the individual points and to use the enormous space and degrees of
freedom that is available in n-dimension space to separate each point,
rather than try to separate clusters which will never be well defined.
The above arguments was the genesis of the idea that gave an impetus
to do the kind of research work in {[}1-3,12{]} and that which is
being presently reported in this paper. This direction has been greatly
encouraged by the recent discovery of an algorithm which enables one
to separate any number of given points in large dimension space in
such a manner that every point is separated from every other by at least
one plane. The best part of this algorithm is that it is non-iterative
and finds the coefficients of the equations of each one of the planes
that separate a given set of points; with a computational complexity
of approx.

$O(n.Nlog(N))+O(n^{3}log(N))$, where $N$ is the given number of
points and $n$ is the dimension of space. (The steps of the algorithm
and its detailed proof are given in {[}1-3,12{]}).

The subject of this paper is pattern recognition and the purpose of
this paper is to show that by means of the above newly discovered
algorithm one can build a noniterative classifier which can assign
sample points in data to an appropriate (correct) class. Actually,
as stated in the abstract, there are two methods which can be used
for the classification of a new test-sample (point): (i) The first
directly uses the information obtained from the planes that separate
the training-sample points each of which are assumed to have a class
label and then tries to classify the new test-sample. (ii) The second
method is based on the first, but uses another {}``Cluster Discovery''
algorithm to find the clusters which belong to each class in the training-sample
feature space and then proceeds to perform a classification by trying
to find out the cluster that the test-point would most probably belong
to. The second method is more like the traditional neural method,
but it is non-iterative and deterministic. The first method would
seemingly be like a nearest neighbour algorithm, but this is not so
as this method determines the exact quadrant to which a neighbouring
point belongs to and not just the distance. As will be clear (later
on) in $n-$dimension space there are $2^{n}$quadrants surrounding
any neighbouring point, (which is a large number), in the present
method since we have separated each point by planes we have in a manner
taken the quadrant information into account, whereas a nearest neighbour
method loses all quadrant and directional information, by reducing
everything to a single number: the distance. It is this aspect that
makes the present method superior as actual application reveals.

The next two sections describe both the methods and the third section
after this gives the details of their application to the well known
IRIS data set.

We delve into both the methods without further ado.

\section{Method 1: A neural classifier by using separation of individual points}

In this method the process is as follows: 

(i) All the data which consists of points in n-dimensional sample
space is divided into two sets (a) a set of points,say $N$ in number
which will be considered as the {}``Traning Set'' and (b) another
set of points which will be considered as the {}``Test Set''. It
will be assumed that each point in the Training set belongs to some
class which is known. The points in the Test Set are assumed to be
belonging to some unknown class which has to be discovered by the
classifier.

(ii) The new algorithm is then used to examine the coordinates of
each point in the Training Set and then to find the coordinates of
the planes that would separate each of the $N$ points in this set
so that each point is separated from the other by atleast one plane,
let q be the number of planes that separate these points.\footnote{The number of planes,$q,$ required to do this would be such that
$q=O(log_{2}(N))$ this empirical estimate of $q$ gets better when
the number of dimensions, $n,$ gets bigger. (For eg. for a particular
set of random points, generated by the authors, where $N=2,000$ and
$n=15$, it was only 22 planes (i.e. $q=22)$ where as for another
case when $N=50,000$ and $n=25$ it was found that only 5 more planes
ie 27 planes$(q=27)$ separated all the points, this increase is as
per the empirical formula since $log_{2}(50,000/2000)=log_{2}(25)=4.65\approx5)$
planes.)See Ref{[}12{]} for details.%
} 

(iii) Once all the q planes and their equations have been found out
the Orientation Vector (OV) of each of the N points is found. The
OV is a Hamming vector, of dimension $q$, which is associated with
each point P belonging to the Training Set. If P is a point in the
Training Set then its orientation vector $Ov(P)$ indicates the {}``orientation''
of the point P w.r.t. each of the $q$ planes See Figure below.
\end{multicols}

\includegraphics[scale=0.4]{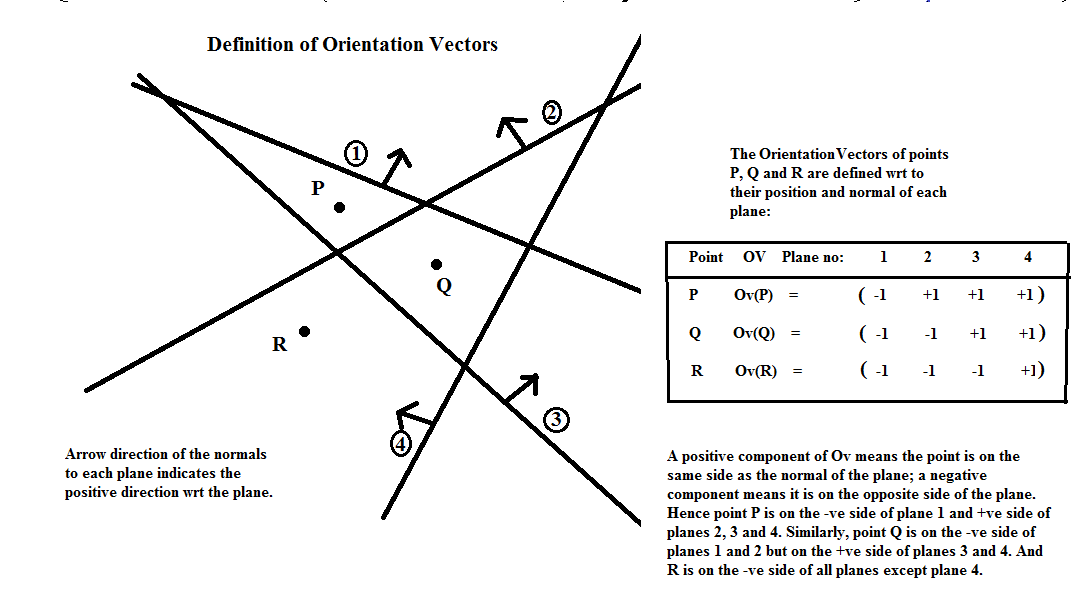}

The Orientation Vector%
\footnote{It is assumed that every plane will have a specific (defined) normal
direction, a point which lies on the positive side of the normal is
said to lie on the positive side of the plane, on the other hand if
the point lies on the other side it is said to lie on the negative
side.This direction is easily found if the equation of the plane is
known for example if $1+\alpha_{1}x_{1}+\alpha_{2}x_{2}+...,+\alpha_{n}x_{n}=0$
is the equation to a particular $n$- dimensional plane then a point
P whose coordinates are $(p_{1,}p_{2},....,p_{n})$ and is not on
the plane, will be said to be on the positive side of this plane if
$1+\alpha_{1}p_{1}+\alpha_{2}p_{2}+...,+\alpha_{n}p_{n}>0$ and in
the negative side if $1+\alpha_{1}p_{1}+\alpha_{2}p_{2}+...,+\alpha_{n}p_{n}<0$.%
} has the property that two points (say) $P$ and $R$ will \textit{not
}have a plane between them\textit{ if and only if }$Ov(P)=Ov(R)$.
That is $Ov(P)\ne Ov(R)$ automatically implies that P and R are separated
by at least one of the $q$ planes which separate each of the points
in the Training Set.\footnote{Another fact, that only adds to the prospect of success in this new direction is that: for large n dimension space where $2^n > N$ (N being the number of points), one will find that the number of planes q,
needed for separating N points is far less than the number of points
itself, in fact q = O(log2(N)).%
}

\begin{multicols}{2}
The algorithm that finds the planes which separate all the points
is given in ref{[}2 ,3{]} and will not be repeated here.

After the points are separated and the planes that have performed
the separation are known then the method to process the classification
of new points is easily devised as we shall see. We use the information
obtained from the planes that separate the training-sample points
each of which are assumed to have a class label and then try to
classify the new test-sample.

\subsection{An application to Data Classification and Retrieval}

We will now briefly describe how this algorithm can be applied to
data retrieval. Suppose one has $N$ points in a $n$ dimension X-Space
and the algorithm was used to separate all N points and it was found
after running the algorithm that $q$ planes were finally necessary.
We will now show that this information can be used as a data retrieval
device. That is, all the data can now be stored in such a manner that
retrieval can be done with great efficiency. 

Now for purposes of this illustration we will assume that each point
N represents one $n$ dimensional sample in X-Space. This sample,
data point, will have $n$ numbers which may relate to a medical data
of one person or alternatively it could be an image involving n pixels
and represent a photograph of a person.Now we wish to store many such
samples, data points, in such a manner that classification and retrieval
becomes easy. The idea is simple: after we had run the algorithm which
separates each of the N data points (samples) by  q number of planes,
we will have the exact information of how each of the data points
$N$ reside in X-Space with respect to each other and the separation
planes, because we have the Orientation Vector for each point stored
in S. We can use this information to (i) store and classify the data
in such a manner that it is possible to (ii) retrieve it easily. What
we mean is that after the storage is done and if we are given a fresh
(approximate) data of a person whose data is stored in the storage
receptacle in the repository, it possible to use this new (approximate)
data to retrieve the stored  data (image). In other words if we are given
a new sample point, we can retrieve another examplar which is closest
to the present sample point. Example if the new sample point is P
and it may represent the medical data of a person P, then we can retrieve
another data point of a person L (stored in the data base), which
is closest to the present sample point P.

We show that if this new sample point P in X-space, is close to some
other sample point L (say) stored in the receptacle then we can discover
this fact by calculating $Ov(P)$, the Orientation Vector of P and
taking a {}``dot product'' with the Orientation Vector $Ov(Q)$,
of all other points Q stored in the data base. Then by comparison
one will find a point L such that the dot product $Ov(P).Ov(L)$ is
a maximum. Oviously then the point L is the one closest to the sample
point P and hence in all probability points P and L will belong to
the same class. This is how we associate the class of the new sample
P with the class of point L, the latter being known. The above steps
can be represented as a diagram which is nothing but a neural engine
in the picture below:
\end{multicols}

\includegraphics[scale=0.40]{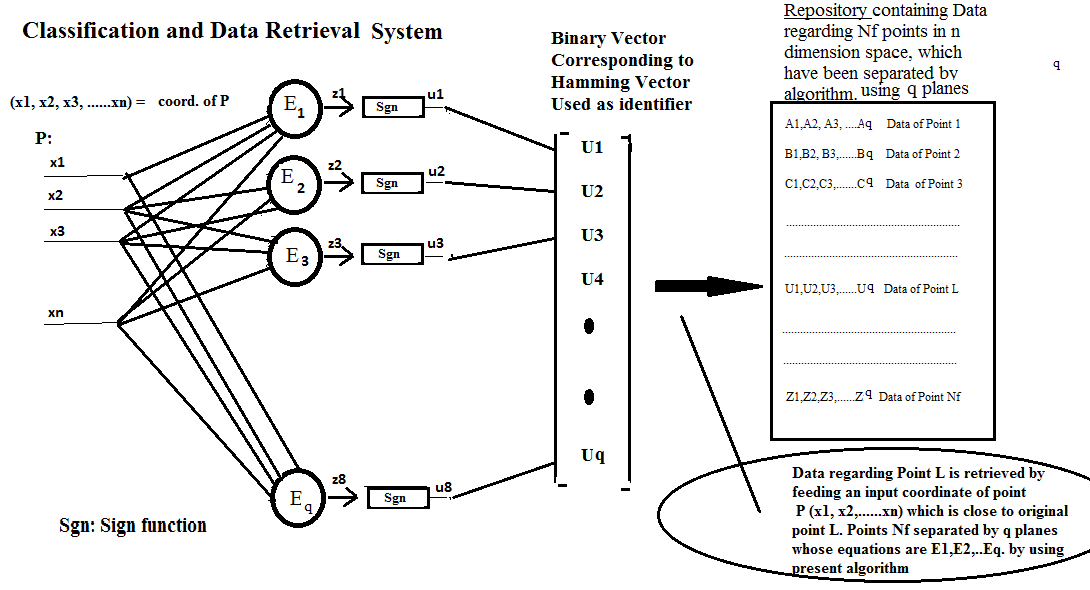}

\begin{multicols}{2}

In the figure $E_{1}$ represent the equation to the first plane stored
in S; We assume that the equation to this plane is given by 
\begin{equation}
1+\alpha_{1}x_{1}+\alpha_{2}x_{2}+...,+\alpha_{n}x_{n}=0
\end{equation}
 we then define 
\begin{equation}
z_{1}=1+\alpha_{1}x_{1}+\alpha_{2}x_{2}+...,+\alpha_{n}x_{n}
\end{equation}
 If we use the Sgn (Sign) function defined as: $Sgn(z)$ defined s.t.
$Sgn(z)=1$, if $z > 0$ and $Sgn(z)=-1$ if $z < 0$ and define $U_{1}=Sgn(z_{1})$,
then $U_{1}$ is 1 if the point P is on the +ve side of plane 1 and
is -1 if it is on the -ve side of plane 1. the same goes for the other
planes $E_{2},E_{3},..E_{q}$. So the output array $(U_{1},U_{2},U_{3},...U_{q})$
is nothing but a Hamming Vector which is the orientation vector $Ov(P)$. Therefore
our {}``Storage Plan'' for each point L in is to use the Hamming Vector 
which is the Orientation Vector $Ov(L)$ of a point $L$ as its label (like a binary label). This label is  like a  pointer to the information about L stored in a receptacle in the space next to this label of L for easy retrieval.

{}``Retrieval Plan'': Present an approximate image of L say P to
the network. It is then possible to immediately retrieve the information
about L.

\subsection{Results on the IRIS FLOWER Classification Problem using Method 1.}

In this sub-section we describe how the classification-retrieval system
can be applied on a typical data. Here we have chosen as our example
the IRIS data set (see Appendix for data) that was first introduced
by Ronald Fisher {[}15{]} in his famous paper which started the field
of multivariate analysis and statistical classification. The reason
which prompted us to choose this particular data set is two fold:
(i) It is a data set which is not too elementary nor too complicated
for us to be drowned in the details and forget the essence of the
new method that we wish to introduce (ii) It is familiar to almost
every researcher across disciplines hence we feel this example is
an appropriate choice which will interest more people to use this new
techique. 

The IRIS data set {[}15-17{]} is a data set which was first collected
by Edgar Anderson and has data regarding three species of flowers:
Iris setosa, Iris virginica and Iris versicolor data. Four features
were measured from each sample: the length and the width of the sepals
and petals, in centimetres, thus the data is 4-dimensional. Based
on the data the task is to classify any other sample flower into one
of the three classes if we are just given these 4 measurements of
that sample (i.e. we try to find the species of the given sample flower).
The data consists of 4 numbers each for 150 flowers (50 samples for
each class). 

We applied our method on the IRIS data set in the following manner:

(i) we separated the data into two sets (a) the training set and (b)
the test set. The test set was obtained from the IRIS data by choosing
every 5th flower, so there were 120 flowers left in the training set
and 29 in test set (it was discovered that two flowers had the same
dimensions so one of them had to be removed, leaving 149 samples).

(ii) We only used the training set of 120 samples, to build our classification
- retrieval system.

\medskip{}

\textbf{Training Process: }

\medskip{}

The classification system was built in the following steps:

Step 1: Use the separation algorithm to separate the 120 samples each
of which can be represented by a point in 4-dimension space, by planes.

\qquad{}To start the algorithm we initially chose in set S, 4 planes
whose equations were of the form: -$\bar{[\bar{x}_{j}]^{-1}x_{j}}+1=0\:\;(j=1\: to\:4)$
where $\bar{x_{j}}$ is the average value of $x_{j}$ of the data
set.

Step 2: We run the algorithm which transfers points from initial set
G to set S such that each point is in its own 'quadrant' drawing planes
when ever necessary . It was found that 24 planes separate all the
120 points. 

Step 3 : By the means of the coefficients of the 24 planes the classification
system shown in the figure can be built, this system basically finds
the OV's of any incoming 4 dimensional point P. In our case there
will be 24 processing elements $E_{1},E_{2},.....,E_{24}$ in the first
layer because there are 24 planes. 

\medskip{}

\textbf{Testing Process:}

\medskip{}

After this for the testing process each of the 29 points were presented
and the OV of each point was found, then the OV's of each point was
compared to the OV's of the other 120 points and the closest match
was found. By this manner:

It was found that: (i) 9 points (point numbers 5, 10, 25, 30, 35,
40, 70, 85 and 110) from the test data had the same OV as some point
in the train data (ii) 9 points (point numbers 15, 20, 45, 50, 75,
100, 105, 140 and 145) from the test data had their OV's differing
from their closest match with the train data by only one component
(iii) 8 points (point numbers 55, 60, 65, 80, 95, 120, 125 and 130)
from the test data had their OV's differing from their closest match
with the train data by only two components (iv) 2 points (point numbers
90 and 115) from the test data had their OV's differing from their
closest match with the train data by only three components (v) 1 point
(point number 135) is wrongly classified as class1 and class2 .

\textbf{Conclusion}: In method 1, only one point (point number 135
was misclassified) out of 29 test points all the others were correctly
classified according to their nearest match (quadrants). Of course since
we are using planes which separate all the 120 training points all
the 120 train points are, obviosly, exactly classified.

\medskip{}

\section{Method 2: Classification by using a Cluster Discovery algorithm in
addition to the point separation method.}

In this method we use the information of the OV's by application of
the previous algorithm, to form pure clusters of the training data,
in such a manner that each cluster has within itself only points belonging
to the same class. This is done by another \textquotedblleft{}cluster
discovery\textquotedblright{} algorithm which is of complexity $O(N^{2}).$
After which it is easily possible to determine the architecture of
a three layer neural network of processing elements and write down the weights
of each processing elements without any more ado. It may be recalled that the weights of the processing elements are nothing but the coefficients of the planes that separate the clusters, since these are known the weights are known. Thus we have  obtained a non-iterative and deterministic method of building a neural classifier.

The Cluster Discovery Algorithm: This is a simple algorithm for obtaining
{}``pure bubbles'' from data. What we mean as a pure bubble is a
spherical region containing only points belonging to one class, if
we are given as input a collection of points the algorithm finds the
{}``bubbles'' which separate points belonging to one class from
points belonging to a different class.
\end{multicols}
\begin{center}
\includegraphics[scale=0.4]{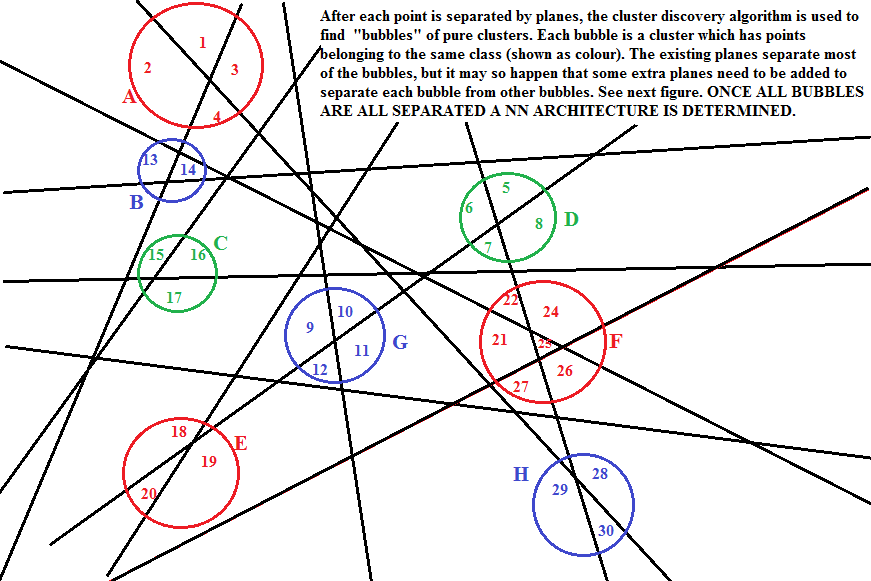}
\end{center}
\begin{multicols}{2}
We are given a set of data points in G belonging to various classes.

Step 1: Begin: Pick a point at random from set G (say) point P, put
it in a set called bubble $B_{P}$ if it belongs to class say class
A, then call the coordinate of P as Center, go to Step 2

Step 2: Then find another nearest point to Center, say Q, if it also
belongs to class A then include Q in $B_{P}$ . Find the {}``radius
$r$ of the bubble as the distance from the Center to its furtherest
point contained in the Bubble. Go to Step 3 if the latest point Q
has the same class as P else go to Step 4

Step 3: Find the centroid of points in Bubble $B_{P}$ call the centroid
a Center then go to Step 2 to find another Q

Step 4 (you will come here when the point Q does not belong to the
class of point P.) Find the radius r between Center and Q, put $r_{j}$,
as $r_{j}=r-\epsilon$ , where $\epsilon$ is a small positive number.(Also check if all points 
inside this radius belong to the same class, else step back and choose the previous bubble.) Give
a name to the bubble $B_{P}$ as say $B_{j}$ with radius $r_{j}$.
Store$B_{j}$ and remove the points inside the bubble $B_{j}$ from
set G. If G is empty Quit else go to Step 1 to Begin again and find
a new bubble around a new point P.

This algorithm finds bubbles (spherical clusters) of various sizes
each bubble having points belonging to the same class and a known
radius.

Once all the points are separated by planes. This algorithm will discover
the small pure clusters (bubbles), now most of the bubbles will be
separated from one another by the existing planes. However it may
some times be required to add additional planes so that two bubbles
are separated by  at least one plane. These additional planes are
shown in red in the figure below. A simple algorithm can be used
to separate two bubbles (say bubble A from B) by drawing a plane perpendicular
to line joining the centers of A and B, the position of this plane
can be suitably chosen to be outside the two bubbles. The OVs of each
point can be used to discover which particular bubbles need to be
separated by additional planes.(Se Figure below:)
\end{multicols}
\begin{center}
\includegraphics[scale=0.4]{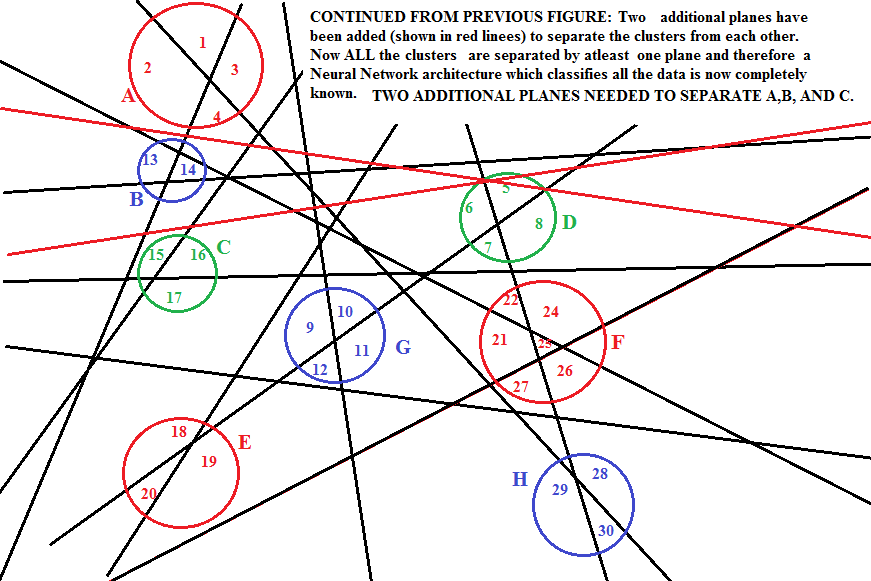}
\end{center}
\begin{multicols}{2}
\textbf{NOTE}: In order to check if the clusters are separable  by planes we need to calculate the cluster OV's, to do this we adopt the same
procedure as that of points, that is if a bubble is to the +ve side
of plane no. 5 (say) its 5th component will be +1 and if it is on
the negative side of plane 5 it will be -1. However if a plane passes
through the bubble then its component is made 0, example if plane
7 passes through the bubble the 7th component is made 0, when this
happens we say that the plane 7 is \textbf{disabled} for that cluster. The cluster OV is the common OV of the points in the cluster after disabling the planes that pass through the cluster.
The procedure is adopted to quickly find if every two bubbles say
A and B are separated from one and another a requirement which will
be met if atleast one plane (which is not disabled wrt to A and B)
separates the two clusters. (Of course while separating points by
planes no planes are disabled because it is ensured that no plane
passes through a point, but this is not guaranteed for extended objects
like clusters).

By using the above method we are able to obtain planes which separate
clusters after they have been discovered by the {}``Cluster Discovery''
algorithm. By using the methods described in Ref {[}18{]} we will
be determine the neural architecture which can classify the clusters.
But very importantly in our case the coefficients of all the planes
are already known, hence the weights of the processing elements are
already determined. Hence this method is non iterative, just like
Method 1.

\subsection{Results on the IRIS FLOWER Classification Problem using Method 2}

We applied the above method for the same IRIS Flower data. And did
as folows: 
\end{multicols}
\begin{center}
\includegraphics[scale=0.50]{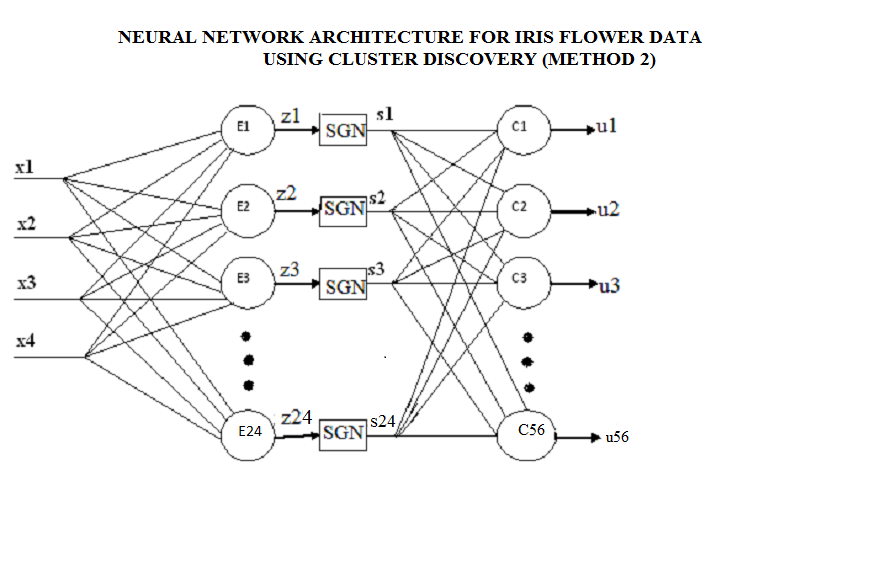}
\end{center}
\begin{multicols}{2}
In this case it was not found necessary to introduce more planes to separate the clusters, 
24 planes were sufficient. And the final group of clusters were 56.

In the figure E1 represents the equation to the first plane stored
in S; we assume that the equation to this plane is given by 
 $1+ \alpha_1 x_1 + \alpha_2 x_2 + \alpha_3 x_3 + \alpha_4 x_4 =0$
 
We then define $z_{1}=1+ 1+ \alpha_1 x_1 + \alpha_2 x_2 + \alpha_3 x_3 + \alpha_4 x_4$
If we use the Sgn (Sign) function defined as: $Sgn(z)$ defined s.t.
$Sgn(z)=1$, if $z>0$ and $Sgn(z)=-1$ if $z < 0$ and define $s_{1}=Sgn(z_{1})$,
then $s_{1}$ is 1 if the point P is on the +ve side of plane 1 and
is -1 if it is on the -ve side of plane 1. The same goes for the other
planes$E_{1},E_{2},E_{3}...,E_{24}$ We examine all the out puts $u_{1},u_{2},u_{3}...,u_{56}$
 obtained from the orientation vector Ov(P) (since by definition $u_j= Ov(P).Ov(c_j)$) and we choose that $u_{j}$ which has
the maximum positive value (dot product). Supose $u_{j}$ is maximum
then cluster $c_{j}$ is considered as the {}``attractor'' and the
input point P, whose coordinates are $(x_{1},x_{2}.x_{3},x_{4})$
is considered to belong to that class which $c_{j}$ belongs. In this
manner all the 29 test points were classified.

\textbf{Procedure and Results:} Initially 86 bubbles were discovered, it was found that by
regrouping bubbles which have the same OVs and which belong to same class
(after disabling of the planes which passes through the bubbles) there
were 56 cluster groups. These cluster OVs were computed by disabling of the planes
which pass through the merged bubbles. All cluster OVs were then unique, though clusters which belong to the same class are
permitted to have the same OVs. As stated above this process of creating bubbles and merging them (discovering clusters), can be easily programmed, using the information of the OVs of all the points in the Training data set. When the OVs of the bubbles belonging to the different classes have the
same OV, we can separate these bubbles by drawing necessary planes.In this particular IRIS case it was not needed to draw any more planes. 
At times when a  point belonging to
one class comes within a central region of a bubble belonging to a different class, 
we need to separate these points by creating new bubbles. It may then happen that there will be bubbles with only one point.
The cluster discovery algorithm does all this automatically.

The crucial and essential point to be  remembered is that the procedures described in this paper is non-iterative and leads to a NN architecture which classifies the data.

 We tested the data on the  neural network shown in figure, by first applying it on 120
training points. All 120 points were correctly classified. We also tested this network 
by passing each of the 29 test points as an input to the NN. Out of 29 points only one 
point was wrongly classified as class 2  (no. 130). All the other 28 points were correctly classified.

\section{Conclusion}

In this paper we have used a new algorithm which separates points
in n-dimension space by planes to develop noniterative classification
techniques for solving pattern recognition problems using neural networks.
The methods developed in this paper are very new and will find application
in neural research and Deep Learning not only because of their non-iterative
and deterministic nature but also because of their efficiency and
speed. It is strongly felt that they have the potential to supersede
other classification methods which are iterative in nature and rely
on error minimization.

\end{multicols}
\section{Acknowledgements}
The authors thank the management of Sreenidhi: Dr P. Narsimaha Reddy and Dr K.T Mahi for their sustained
support; the HOD Dr Aruna Varanasi and their colleagues for creating a happy environment. 
Both the authors K.E. and K.D.R, thank their spouses Suhasini and
Amani, resp. for their unwavering faith and encouragement. 

\begin{small}
\section{References}

{[}1{]} K.Eswaran: A system and method of classification etc. Patents
filed IPO No.(a) 1256/CHE July 2006 and (b) 2669/CHE June 2015

{[}2{]} K.Eswaran: On a non-iterative algorithm for separation of
points in n-dimension space by planes , 2015 (sent for publication):
arxiv.orgslash-abs-slash1509.08742

{[}3{]} K.Eswaran : On the storage and retrieval of primes and otheer
random numbers using n-dimensional geometry, 2015, (sent for publication):
arxiv.org-slash-abs-slash-1511.08941

{[}4{]} R.A. Fischer: The statistical utilization of multiple measurements,
Annals of Eugenics, 8, 376-386 (1938); also Annals Eugenics, 7, 179-188
(1936) 

{[}5{]} P.C. Mahalanobis: On the generalized distance in statistics,
Proc. Nat. Inst. of Sc. India, 12, 49-55 (1936)

{[}6{]} McCullough, W. and Pitts, W. A: Logical calculus of the ideas
immanent in nervous activity, Bulletin of Mathematical Biophysics,
5:115133. (1943)

{[}7{]} A. N. Kolmogorov: On the representation of continuous functions
of many variables by superpositions of continuous functions of one
variable and addition. 

Doklay Akademii Nauk USSR, 14(5):953 - 956,(1957). Translated in:
Amer. Math Soc. Transl. 28, 55-59 (1963).

{[}8{]} PaulWerbos: Beyond Regression: New Tools for Prediction and
Analysis in the Behavioral Sciences, PhD thesis, Harvard University,
1974 

{[}9{]} Rumelhart, D. E., Hinton, G. E., and R. J. Williams: Learning
representations by back-propagating errors. Nature, 323, 533536 (1986).

{[}10{]} J. Schmidhuber: Deep Learning in Neural Networks: An Overview.
75 pages, www.arxiv.org-slashabs-slash1404.7828 (2014). 

{[}11{]}Yoshua Bengio: Learning Deep Architectures for AI. Foundations
and Trends in Machine Learning: Vol. 2: No. 1, pp 1-127 (2009). 7

{[}12{]} K. Eswaran: Results of Separation of points in high dimension
space from a computer program based on the non-iterative algorithm,

$\;\;$Tech Report: www.researchgate.net(slash)publication(slash)285149835
(slash)Results

{[}13{]} J.C. Hawkins with S. Blakeslee: On Intelligence, Publ. Henry
Holt and Co. NY. (2004) 

{[}14{]}D. George and J.C. Hawkins: Trainable hierarchical memory
system and method, January 24 2012.

URL www.google.com patents US8103603. US Patent 8,103,603 

{[}15{]} R. A. Fisher (1936). ``The use of multiple measurements
in taxonomic problems'' (PDF). Annals of Eugenics 7 (2):
179-188. doi:10.1111/j.1469-1809.1936.tb02137.x. 

{[}16{]} Edgar Anderson (1936). ``The species problem
in Iris''. Annals of the Missouri Botanical Garden 23
(3): 457-509. JSTOR 2394164. 

http: archive.ics.uci.edu-slash-ml-slash-machine-learning-databases-slash-iris

{[}17{]} Edgar Anderson (1935). ``The irises of the Gaspé
Peninsula''. Bulletin of the American Iris Society 59:
2-5.

{[}18{]} K Eswaran and Vishwajeet Singh. Article: Some Theorems for
Feed Forward Neural Networks. International Journal of Computer Applications
130(7):1-17, November 2015. 

\medskip{}

\end{small}

\section{APPENDIX: PARTICULARS OF IRIS DATA AND PLANE COEFFICIENTS}

Iris data were taken from http: archive.ics.uci.edu (slash)ml(slash)machine-learning-databases(slash)
iris. Points 102 and 143 had the same coordinates, so we removed point 143. The order of
points from 1 to 142 were same as before, but the points 144 to point 150 were renumbered as 143 to 149.
\medskip{}
\textbf{Results of Method 1:}
\medskip{}
For testing we used 29 points which are points: 5,10,15,20,25,30,35,40,45,50,55,60,65,70,75,80,85,90,95,100,105,110,115,120,125,130,135,140 and 145 (ie is we just took every 5th point as a test point to avoid bias).

For training we used 120 points which were the points other than testing points.

\medskip{}

All the equations of the planes are of the form:
$1+ \alpha_1 x_1 + \alpha_2 x_2 + \alpha_3 x_3 + \alpha_4 x_4 =0$.

The coefficients $\alpha_1, \alpha_2, \alpha3, \alpha_4$ 
which were found by the separation algorithm {[}2-3{]} are given below for
all the 24 planes:
\medskip{}

1 -0.171135196 0.0 0.0 0.0 

2 0.0 -0.327082424 0.0 0.0 

3 0.0 0.0 -0.266098988 0.0 

4 0.0 0.0 0.0 -0.833796553

5 0.0915693084 -0.179981054 -0.982001894 2.39974739

6 -3.01654425 3.39045923 3.27472364 -5.11462274 

7 -1.06402164 -1.24436429 1.63510670 1.03997595 

8 -0.03822102608 -0.121900195 -0.0407087851 -0.151979464 

9 0.0229477417 0.0500678001 -0.281631375 -0.00834463335 

10 -0.170647862 -0.156544733 0.152666372 -0.165359189 

11 0.00753583558 -0.60146733 -0.0748200819 0.610617992 

12 0.161856393 -0.373535929 -0.104325804 -0.221864376 

13 0.183604336 -0.987804878 -0.377032520 1.31029810 

14 -0.259943020 0.0116093648 0.270751739 -0.411665410 

15 -0.288949841 -0.187926024 0.272112538 -0.00434511039 

16 -0.135210559 -0.131751684 0.0658758421 -0.0248410096 

17 -0.524137931 0.758620689 -0.317241379 -2.55172413 

18 -0.238095238 0.384615384 -0.540293040 -1.02564102 

19 -0.0861793807 -0.0127673156 -0.370252154 0.194701563 

20 -0.952380952 2.85714285 -4.76190476 1.90476190 

21 -2.39533233 2.68548450 2.73752266 -4.10470708 

22 -0.564408491 0.208095213 0.368554173 0.0940926620 

23 -0.0959491471 -0.0793593893 -0.390971649 0.859509858 

24 -0.585521309 0.434385125 0.397997055 -0.335871170 

\medskip{}

\textbf{Results of Method 2:}
\medskip{}

\textbf{(i) List of Bubbles} 
\medskip{}

The following list contains the points within each bubble. The suffixes indicate the bubble number of the bubble. For example bubble number 5, contains points:22,47,49. THE TOTAL NUMBER OF BUBBLES = 86.
\medskip{}

$(1,18)_1,(6,9,11)_2,(16,34,17)_3,(21,32)_4,(22,47,49)_5,(33)_6,(37)_7,(41)_8,(14)_9,(42)_{10},(9)_{11},(39)_{12},$
$(43)_{13},(4)_{14},(48)_{15},(13)_{16},(3)_{17},(46)_{18},(7)_{19},(2)_{20},(23)_{21},(31)_{22},(26)_{23},(36)_{24},$
$(12)_{25},(8)_{26},(38)_{27},(27)_{28},(24)_{29},(44)_{30},(29)_{31},(28)_{32},(51,53)_{33},(52,57,92)_{34},(54,81,82)_{35},$
$(56,67,97)_{36},(59,76,66)_{37},(61,94)_{38},(62)_{39},(63,93,68)_{40},(64)_{41},(69,88,73)_{42},(71)_{43},(72,98)_{44},$
$(74)_{45},(77)_{46},(78)_{47},(79)_{48},(83)_{49},(84)_{50},(86)_{51},(89,96)_{52},(91)_{53},(99)_{54},(87)_{55},(58)_{56},$
$(101,137)_{57},(102,114,122)_{58},(103)_{59},(104,117,138)_{60},(106,123,119)_{61},(108)_{62},(109,129,133)_{63},$
$(111)_{64},(112)_{65},(113)_{66},(118,132)_{67},(124)_{68},(134)_{69},(136)_{70},(139)_{71},(142)_{72},(143)_{73},$
$(146)_{74},(148)_{75},(149)_{76},(144)_{77},(107)_{78},(126)_{79},(131)_{80},(147)_{81},(116)_{82},(141)_{83},$
$(121)_{84},(127)_{85},(128)_{86} $
\medskip{}

\textbf{(ii) List of Clusters}
\medskip{}
  The 86  bubbles were then merged/ regrouped to 56 clusters. All cluster OVs were then unique, though clusters which belong to the same class are
permitted to have the same OVs. In this particular case it was not required to introduce any additional planes to separate the bubbles which belong to different classes. The distribution of the 86 bubbles into 56 clusters are given below:
\medskip{}

Cluster 1 contains Bubbles: (1,2,3,4,5,6,8,26,27,28,31,32); Cluster 2: (7); Cluster 3:(9); 4:(10); 5:(11); 6:(12); 7:(13); 8:(14); 9:(15); 10:(16); 11:(17); 12:(18); 13:(19); 14:(20); 15:(21); 16:(22); 17:(23); 18:(24); 19:(25); 20:(29); 21:(30); 22:(33,34,37,39,44,48,51,46,55); 23:(35,38,40,42); 24:(36,52,53); 25:(41); 26:(43); 27:(45); 28:(47); 29:(49); 30:(50); 31:(54); 32:(56); 33:(63,65); 34:(81,82,83,84,85,86); 35:(57); 36:(58); 37:(59); 38:(60); 39:(61); 40:(62); 41:(67); 42:(69); 43:(70); 44:(71); 45:(72); 46:(73); 47:(74); 48:(75); 49:(76); 50:(78); 51:(79); 52:(80); 53:(64); 54:(66); 55:(68) and Cluster 56:(77).
\medskip{}

The numbers calculated by the two algorithms and given in this Appendix   completely determined the NN architecture for the IRIS problem as shown in the last section of the paper and with results described therein.
\medskip{}

------------------

SUPPLEMENTARY MATERIAL
\medskip{}

(i) Training points of the IRIS FLOWER DATA SET
The coordinates of all the 120 Training points are given below.
Note every 5th Point was taken to be test point, their data was not taken into account for training. The 24 planes whose coefficients are given in the Appendix separate only these 120 points.

$PointNo     x_1, x_2, x_3, x_4$ 

1 5.1  3.5  1.4  0.2
 
2 4.9  3.0  1.4  0.2 

3 4.7  3.2  1.3  0.2 

4 4.6  3.1  1.5  0.2 

6 5.4  3.9  1.7  0.4 

7 4.6  3.4  1.4  0.3 

8 5.0  3.4  1.5  0.2 

9 4.4  2.9  1.4  0.2 

11 5.4  3.7  1.5  0.2
 
12 4.8  3.4  1.6  0.2 

13 4.8  3.0  1.4  0.1 

14 4.3  3.0  1.1  0.1 

16 5.7  4.4  1.5  0.4 

17 5.4  3.9  1.3  0.4 

18 5.1  3.5  1.4  0.3 

19 5.7  3.8  1.7  0.3 

21 5.4  3.4  1.7  0.2 

22 5.1  3.7  1.5  0.4 

23 4.6  3.6  1.0  0.2 

24 5.1  3.3  1.7  0.5 

26 5.0  3.0  1.6  0.2 

27 5.0  3.4  1.6  0.4 

28 5.2  3.5  1.5  0.2 

29 5.2  3.4  1.4  0.2 

31 4.8  3.1  1.6  0.2 

32 5.4  3.4  1.5  0.4 

33 5.2  4.1  1.5  0.1 

34 5.5  4.2  1.4  0.2 

36 5.0  3.2  1.2  0.2 

37 5.5  3.5  1.3  0.2 

38 4.9  3.6  1.4  0.1
 
39 4.4  3.0  1.3  0.2 

41 5.0  3.5  1.3  0.3 

42 4.5  2.3  1.3  0.3 

43 4.4  3.2  1.3  0.2 

44 5.0  3.5  1.6  0.6 

46 4.8  3.0  1.4  0.3 

47 5.1  3.8  1.6  0.2 

48 4.6  3.2  1.4  0.2 

49 5.3  3.7  1.5  0.2 

51 7.0  3.2  4.7  1.4 

52 6.4  3.2  4.5  1.5
 
53 6.9  3.1  4.9  1.5 

54 5.5  2.3  4.0  1.3 

56 5.7  2.8  4.5  1.3 

57 6.3  3.3  4.7  1.6 

58 4.9  2.4  3.3  1.0 

59 6.6  2.9  4.6  1.3 

61 5.0  2.0  3.5  1.0 

62 5.9  3.0  4.2  1.5 

63 6.0  2.2  4.0  1.0 

64 6.1  2.9  4.7  1.4 

66 6.7  3.1  4.4  1.4 

67 5.6  3.0  4.5  1.5 

68 5.8  2.7  4.1  1.0 

69 6.2  2.2  4.5  1.5 

71 5.9  3.2  4.8  1.8 

72 6.1  2.8  4.0  1.3 

73 6.3  2.5  4.9  1.5 

74 6.1  2.8  4.7  1.2 

76 6.6  3.0  4.4  1.4 

77 6.8  2.8  4.8  1.4 

78 6.7  3.0  5.0  1.7 

79 6.0  2.9  4.5  1.5 

81 5.5  2.4  3.8  1.1 

82 5.5  2.4  3.7  1.0 

83 5.8  2.7  3.9  1.2 

84 6.0  2.7  5.1  1.6 

86 6.0  3.4  4.5  1.6 

87 6.7  3.1  4.7  1.5 

88 6.3  2.3  4.4  1.3 

89 5.6  3.0  4.1  1.3 

91 5.5  2.6  4.4  1.2 

92 6.1  3.0  4.6  1.4 

93 5.8  2.6  4.0  1.2 

94 5.0  2.3  3.3  1.0 

96 5.7  3.0  4.2  1.2 

97 5.7  2.9  4.2  1.3 

98 6.2  2.9  4.3  1.3 

99 5.1  2.5  3.0  1.1 

101 6.3  3.3  6.0  2.5 

102 5.8  2.7  5.1  1.9 

103 7.1  3.0  5.9  2.1 

104 6.3  2.9  5.6  1.8 

106 7.6  3.0  6.6  2.1 

107 4.9  2.5  4.5  1.7 

108 7.3  2.9  6.3  1.8 

109 6.7  2.5  5.8  1.8 

111 6.5  3.2  5.1  2.0 

112 6.4  2.7  5.3  1.9 

113 6.8  3.0  5.5  2.1 

114 5.7  2.5  5.0  2.0 

116 6.4  3.2  5.3  2.3 

117 6.5  3.0  5.5  1.8 

118 7.7  3.8  6.7  2.2 

119 7.7  2.6  6.9  2.3 

121 6.9  3.2  5.7  2.3 

122 5.6  2.8  4.9  2.0 

123 7.7  2.8  6.7  2.0 

124 6.3  2.7  4.9  1.8 

126 7.2  3.2  6.0  1.8 

127 6.2  2.8  4.8  1.8 

128 6.1  3.0  4.9  1.8 

129 6.4  2.8  5.6  2.1 

131 7.4  2.8  6.1  1.9 

132 7.9  3.8  6.4  2.0 

133 6.4  2.8  5.6  2.2 

134 6.3  2.8  5.1  1.5 

136 7.7  3.0  6.1  2.3 

137 6.3  3.4  5.6  2.4 

138 6.4  3.1  5.5  1.8 

139 6.0  3.0  4.8  1.8 

141 6.7  3.1  5.6  2.4 

142 6.9  3.1  5.1  2.3 

143 6.8  3.2  5.9  2.3 

144 6.7  3.3  5.7  2.5 

146 6.3  2.5  5.0  1.9 

147 6.5  3.0  5.2  2.0 

148 6.2  3.4  5.4  2.3 

149 5.9  3.0  5.1  1.8  

The 24 planes whose coefficients are given in the Appendix separate only these 120 points. The Orientation vectors of the 120 points w.r.t these 24 planes are given below. (See foot note 2 on page 3, to recall how the OVs are defined.)

It may be easily checked that each of the 120 OVs given below is unique, thus proving that the 24 planes really separate each of the the 120 test points from one another.

ORIENTATION VECTORS OF TRAINING POINTS:

OV(  1) =   1 -1  1  1 -1  1 -1  1  1 -1 -1  1 -1  1 -1 -1  1  1  1 -1  1 -1 -1  1
    
OV(  2) =   1  1  1  1  1 -1 -1  1  1 -1 -1  1 -1  1 -1  1 -1  1  1 -1  1 -1 -1 -1
    
OV(  3) =   1 -1  1  1  1  1 -1  1  1 -1 -1  1 -1  1 -1  1  1  1  1 -1  1 -1 -1  1
    
OV(  4) =   1 -1  1  1 -1  1 -1  1  1 -1 -1  1 -1  1 -1  1 -1  1  1 -1  1 -1 -1  1
    
OV(  6) =   1 -1  1  1  1  1 -1  1  1 -1 -1  1 -1 -1 -1 -1 -1 -1 -1 -1  1 -1 -1  1
    
OV(  7) =   1 -1  1  1  1  1 -1  1  1 -1 -1  1 -1  1 -1  1 -1  1  1  1  1 -1 -1  1
    
OV(  8) =   1 -1  1  1 -1  1 -1  1  1 -1 -1  1 -1  1 -1 -1 -1  1  1 -1  1 -1 -1  1
    
OV(  9) =   1  1  1  1 -1  1 -1  1  1 -1 -1  1 -1  1 -1  1 -1  1  1 -1  1 -1 -1  1
    
OV( 11) =   1 -1  1  1 -1  1 -1  1  1 -1 -1  1 -1 -1 -1 -1 -1  1 -1 -1  1 -1 -1 -1
    
OV( 12) =   1 -1  1  1 -1  1 -1  1  1 -1 -1  1 -1  1 -1  1  1  1 -1 -1  1 -1 -1  1
    
OV( 13) =   1  1  1  1 -1  1 -1  1  1 -1 -1  1 -1  1 -1  1  1  1  1 -1  1 -1 -1  1
    
OV( 14) =   1  1  1  1  1  1 -1  1  1 -1 -1  1 -1  1 -1  1  1  1  1  1  1 -1  1  1
    
OV( 16) =   1 -1  1  1  1  1 -1  1  1 -1 -1  1 -1 -1 -1 -1 -1  1 -1  1  1 -1 -1  1
    
OV( 17) =   1 -1  1  1  1  1 -1  1  1 -1 -1  1 -1 -1 -1 -1 -1  1  1  1  1 -1  1 -1
    
OV( 18) =   1 -1  1  1  1  1 -1  1  1 -1 -1  1 -1 -1 -1 -1 -1  1  1  1  1 -1 -1 -1
    
OV( 19) =   1 -1  1  1 -1  1 -1  1  1 -1 -1  1 -1 -1 -1 -1 -1 -1 -1 -1  1 -1 -1 -1
    
OV( 21) =   1 -1  1  1 -1  1 -1  1  1 -1 -1  1 -1  1 -1 -1 -1 -1 -1 -1  1 -1 -1 -1
    
OV( 22) =   1 -1  1  1  1  1 -1  1  1 -1 -1  1 -1 -1 -1 -1 -1 -1  1  1  1 -1 -1  1
    
OV( 23) =   1 -1  1  1  1  1 -1  1  1 -1 -1  1 -1  1 -1 -1  1  1  1  1  1 -1  1  1
    
OV( 24) =   1 -1  1  1  1 -1 -1  1  1 -1 -1  1 -1 -1 -1 -1 -1 -1 -1 -1  1 -1  1 -1
    
OV( 26) =   1  1  1  1 -1  1 -1  1  1 -1 -1  1 -1  1 -1  1 -1 -1 -1 -1  1 -1 -1 -1
    
OV( 27) =   1 -1  1  1  1  1 -1  1  1 -1 -1  1 -1  1 -1 -1 -1 -1  1 -1  1 -1 -1  1
    
OV( 28) =   1 -1  1  1 -1  1 -1  1  1 -1 -1  1 -1  1 -1 -1 -1  1 -1 -1  1 -1 -1  1
    
OV( 29) =   1 -1  1  1 -1  1 -1  1  1 -1 -1  1 -1 -1 -1 -1 -1  1  1 -1  1 -1 -1 -1
    
OV( 31) =   1 -1  1  1 -1  1 -1  1  1 -1 -1  1 -1  1 -1  1 -1 -1 -1 -1  1 -1 -1  1
    
OV( 32) =   1 -1  1  1  1 -1 -1  1  1 -1 -1  1 -1 -1 -1 -1 -1 -1  1 -1 -1 -1 -1 -1
    
OV( 33) =   1 -1  1  1 -1  1 -1  1  1 -1 -1  1 -1  1 -1 -1  1  1 -1  1  1 -1 -1  1
    
OV( 34) =   1 -1  1  1 -1  1 -1  1  1 -1 -1  1 -1 -1 -1 -1  1  1 -1  1  1 -1 -1  1
    
OV( 36) =   1 -1  1  1  1 -1 -1  1  1 -1 -1  1 -1 -1 -1 -1 -1  1  1  1  1 -1 -1 -1
    
OV( 37) =   1 -1  1  1  1 -1 -1  1  1 -1 -1  1 -1 -1 -1 -1 -1  1  1 -1 -1 -1 -1 -1
    
OV( 38) =   1 -1  1  1 -1  1 -1  1  1 -1 -1  1 -1  1 -1 -1  1  1  1  1  1 -1 -1  1
    
OV( 39) =   1  1  1  1  1  1 -1  1  1 -1 -1  1 -1  1 -1  1  1  1  1 -1  1 -1  1  1
    
OV( 41) =   1 -1  1  1  1  1 -1  1  1 -1 -1  1 -1 -1 -1 -1 -1  1  1  1  1 -1 -1  1
    
OV( 42) =   1  1  1  1  1 -1 -1  1  1  1 -1  1 -1  1 -1  1 -1 -1  1 -1 -1 -1  1 -1
    
OV( 43) =   1 -1  1  1  1  1 -1  1  1 -1 -1  1 -1  1 -1  1  1  1  1  1  1 -1 -1  1
    
OV( 44) =   1 -1  1  1  1 -1 -1  1  1 -1 -1  1 -1 -1 -1 -1 -1 -1  1 -1  1 -1  1  1
    
OV( 46) =   1  1  1  1  1 -1 -1  1  1 -1 -1  1 -1  1 -1  1 -1 -1  1 -1  1 -1  1 -1
    
OV( 47) =   1 -1  1  1 -1  1 -1  1  1 -1 -1  1 -1  1 -1 -1  1  1 -1 -1  1 -1 -1  1
    
OV( 48) =   1 -1  1  1 -1  1 -1  1  1 -1 -1  1 -1  1 -1  1  1  1  1 -1  1 -1 -1  1
    
OV( 49) =   1 -1  1  1 -1  1 -1  1  1 -1 -1  1 -1 -1 -1 -1  1  1 -1 -1  1 -1 -1  1
    
OV( 51) =  -1 -1 -1 -1 -1 -1 -1 -1 -1 -1 -1  1 -1 -1 -1 -1 -1 -1 -1 -1 -1 -1 -1 -1
    
OV( 52) =  -1 -1 -1 -1  1 -1 -1 -1  1 -1 -1  1 -1 -1 -1 -1 -1 -1 -1 -1  1 -1 -1 -1
    
OV( 53) =  -1 -1 -1 -1 -1 -1 -1 -1 -1 -1 -1  1 -1 -1 -1 -1 -1 -1 -1 -1  1 -1 -1 -1
    
OV( 54) =   1  1 -1 -1  1 -1  1  1  1  1  1  1 -1  1  1  1 -1 -1 -1 -1 -1 -1 -1 -1
    
OV( 56) =   1  1 -1 -1 -1  1  1  1 -1  1 -1  1 -1  1  1  1 -1 -1 -1 -1  1  1 -1  1
    
OV( 57) =  -1 -1 -1 -1  1  1 -1 -1 -1 -1 -1 -1 -1  1 -1 -1 -1 -1 -1 -1  1  1 -1  1
    
OV( 58) =   1  1  1  1  1  1 -1  1  1  1 -1  1 -1  1  1  1 -1 -1 -1 -1  1  1 -1  1
    
OV( 59) =  -1  1 -1 -1 -1 -1 -1  1 -1 -1 -1  1 -1  1 -1 -1 -1 -1 -1 -1  1 -1 -1 -1
    
OV( 61) =   1  1  1  1  1 -1 -1  1  1  1  1  1 -1  1  1  1 -1 -1 -1 -1 -1 -1 -1 -1
    
OV( 62) =  -1  1 -1 -1  1 -1 -1  1  1 -1 -1  1 -1  1 -1  1 -1 -1 -1 -1  1 -1 -1  1
    
OV( 63) =  -1  1 -1  1 -1 -1 -1  1  1  1  1  1 -1  1 -1  1 -1 -1 -1 -1 -1 -1 -1 -1
    
OV( 64) =  -1  1 -1 -1 -1  1  1  1 -1 -1 -1  1 -1  1 -1  1 -1 -1 -1 -1  1  1 -1  1
    
OV( 66) =  -1 -1 -1 -1  1 -1 -1 -1  1 -1 -1  1 -1 -1 -1 -1 -1 -1 -1 -1 -1 -1 -1 -1
    
OV( 67) =   1  1 -1 -1  1  1  1  1 -1  1 -1 -1 -1  1  1  1 -1 -1 -1 -1  1  1 -1  1
    
OV( 68) =   1  1 -1  1 -1  1 -1  1  1  1 -1  1 -1  1 -1  1 -1 -1 -1 -1  1 -1 -1  1
    
OV( 69) =  -1  1 -1 -1  1 -1  1  1 -1  1  1  1  1  1  1  1 -1 -1 -1 -1 -1 -1 -1 -1
    
OV( 71) =  -1 -1 -1 -1  1  1  1 -1 -1 -1 -1 -1 -1  1 -1  1 -1 -1 -1 -1  1  1 -1  1
    
OV( 72) =  -1  1 -1 -1  1 -1 -1  1  1 -1 -1  1 -1 -1 -1  1 -1 -1 -1 -1 -1 -1 -1 -1
    
OV( 73) =  -1  1 -1 -1 -1 -1  1  1 -1  1  1  1 -1  1  1  1 -1 -1 -1 -1 -1 -1 -1 -1
    
OV( 74) =  -1  1 -1 -1 -1  1 -1  1 -1  1 -1  1 -1  1 -1  1 -1 -1 -1 -1  1 -1 -1  1
    
OV( 76) =  -1  1 -1 -1  1 -1 -1 -1  1 -1 -1  1 -1 -1 -1 -1 -1 -1 -1 -1 -1 -1 -1 -1
    
OV( 77) =  -1  1 -1 -1 -1 -1 -1 -1 -1 -1 -1  1 -1 -1 -1 -1 -1 -1 -1 -1 -1 -1 -1 -1
    
OV( 78) =  -1  1 -1 -1  1 -1  1 -1 -1 -1 -1  1 -1 -1 -1 -1 -1 -1 -1 -1 -1 -1 -1 -1
    
OV( 79) =  -1  1 -1 -1  1 -1 -1  1  1 -1 -1  1 -1  1 -1  1 -1 -1 -1 -1  1  1 -1  1
    
OV( 81) =   1  1 -1  1 -1 -1 -1  1  1  1 -1  1 -1  1 -1  1 -1 -1 -1 -1  1 -1 -1 -1
    
OV( 82) =   1  1  1  1 -1 -1 -1  1  1  1 -1  1 -1  1 -1  1 -1 -1 -1 -1  1 -1 -1 -1
    
OV( 83) =   1  1 -1 -1  1 -1 -1  1  1 -1 -1  1 -1  1 -1  1 -1 -1 -1 -1  1 -1 -1 -1
    
OV( 84) =  -1  1 -1 -1 -1  1  1 -1 -1  1  1  1 -1  1  1  1 -1 -1 -1 -1  1  1 -1  1
    
OV( 86) =  -1 -1 -1 -1  1  1 -1 -1  1 -1 -1 -1 -1  1 -1 -1 -1 -1 -1 -1  1  1 -1  1
    
OV( 87) =  -1 -1 -1 -1  1 -1 -1 -1 -1 -1 -1  1 -1 -1 -1 -1 -1 -1 -1 -1 -1 -1 -1 -1
    
OV( 88) =  -1  1 -1 -1 -1 -1 -1  1  1  1  1  1 -1  1 -1  1 -1 -1 -1 -1 -1 -1 -1 -1
    
OV( 89) =   1  1 -1 -1  1  1 -1  1  1 -1 -1  1 -1  1 -1  1 -1 -1 -1 -1  1  1 -1  1
    
OV( 91) =   1  1 -1 -1 -1  1  1  1  1  1 -1  1 -1  1  1  1 -1 -1 -1 -1  1  1 -1  1
    
OV( 92) =  -1  1 -1 -1 -1  1 -1  1 -1 -1 -1  1 -1  1 -1  1 -1 -1 -1 -1  1  1 -1  1
    
OV( 93) =   1  1 -1 -1  1 -1 -1  1  1  1 -1  1 -1  1 -1  1 -1 -1 -1 -1  1 -1 -1 -1
    
OV( 94) =   1  1  1  1  1 -1 -1  1  1  1  1  1 -1  1  1  1 -1 -1 -1 -1  1 -1 -1  1
    
OV( 96) =   1  1 -1 -1 -1  1 -1  1  1  1 -1  1 -1  1 -1  1 -1 -1 -1 -1  1  1 -1  1
    
OV( 97) =   1  1 -1 -1 -1  1 -1  1  1 -1 -1  1 -1  1 -1  1 -1 -1 -1 -1  1  1 -1  1
    
OV( 98) =  -1  1 -1 -1 -1 -1 -1  1  1 -1 -1  1 -1  1 -1  1 -1 -1 -1 -1  1 -1 -1 -1
    
OV( 99) =   1  1  1  1  1 -1 -1  1  1  1 -1  1 -1  1 -1  1 -1 -1 -1 -1 -1 -1  1 -1
    
OV(101) =  -1 -1 -1 -1  1  1  1 -1 -1 -1  1 -1 -1 -1  1  1 -1 -1 -1 -1  1  1 -1  1
    
OV(102) =   1  1 -1 -1  1 -1  1 -1 -1  1  1 -1 -1  1  1  1 -1 -1 -1 -1  1  1 -1  1
    
OV(103) =  -1  1 -1 -1  1 -1  1 -1 -1 -1  1 -1 -1 -1 -1 -1 -1 -1 -1 -1 -1 -1 -1 -1
    
OV(104) =  -1  1 -1 -1 -1  1  1 -1 -1  1 -1 -1 -1  1  1  1 -1 -1 -1 -1  1  1 -1  1
    
OV(106) =  -1  1 -1 -1 -1 -1  1 -1 -1 -1  1 -1 -1 -1  1 -1 -1 -1 -1 -1  1 -1 -1 -1
    
OV(107) =   1  1 -1 -1  1  1  1  1 -1  1  1  1 -1  1  1  1 -1 -1 -1 -1  1  1  1  1
    
OV(108) =  -1  1 -1 -1 -1  1  1 -1 -1 -1 -1  1 -1  1  1  1 -1 -1 -1 -1  1 -1 -1 -1
    
OV(109) =  -1  1 -1 -1 -1 -1  1 -1 -1  1  1  1 -1  1  1  1 -1 -1 -1 -1  1  1 -1 -1
    
OV(111) =  -1 -1 -1 -1  1 -1  1 -1 -1 -1 -1 -1 -1 -1 -1 -1 -1 -1 -1 -1 -1  1 -1 -1
    
OV(112) =  -1  1 -1 -1  1 -1  1 -1 -1 -1  1  1 -1  1  1  1 -1 -1 -1 -1 -1  1 -1 -1
    
OV(113) =  -1  1 -1 -1  1 -1  1 -1 -1 -1  1 -1 -1 -1 -1 -1 -1 -1 -1 -1 -1  1 -1 -1
    
OV(114) =   1  1 -1 -1  1 -1  1 -1 -1  1  1  1  1  1  1  1 -1 -1 -1 -1 -1  1  1  1
    
OV(116) =  -1 -1 -1 -1  1 -1  1 -1 -1 -1  1 -1  1 -1 -1  1 -1 -1 -1 -1 -1  1  1 -1
    
OV(117) =  -1  1 -1 -1 -1  1  1 -1 -1 -1 -1 -1 -1  1  1  1 -1 -1 -1 -1  1  1 -1  1
    
OV(118) =  -1 -1 -1 -1 -1  1  1 -1 -1 -1 -1 -1 -1 -1 -1 -1 -1 -1 -1 -1  1  1 -1  1
    
OV(119) =  -1  1 -1 -1 -1 -1  1 -1 -1 -1  1  1  1 -1  1  1 -1 -1 -1 -1 -1 -1 -1 -1
    
OV(121) =  -1 -1 -1 -1  1 -1  1 -1 -1 -1  1 -1 -1 -1 -1 -1 -1 -1 -1 -1 -1  1 -1 -1
    
OV(122) =   1  1 -1 -1  1 -1  1 -1 -1  1  1 -1  1  1  1  1 -1 -1 -1 -1  1  1  1  1
    
OV(123) =  -1  1 -1 -1 -1 -1  1 -1 -1 -1  1  1 -1  1  1 -1 -1 -1 -1 -1  1 -1 -1 -1
    
OV(124) =  -1  1 -1 -1  1 -1  1 -1 -1 -1  1  1  1 -1 -1  1 -1 -1 -1 -1 -1 -1 -1 -1
    
OV(126) =  -1 -1 -1 -1 -1  1  1 -1 -1 -1 -1 -1 -1  1 -1 -1 -1 -1 -1 -1  1 -1 -1 -1
    
OV(127) =  -1  1 -1 -1  1 -1  1 -1 -1 -1  1  1 -1 -1 -1  1 -1 -1 -1 -1 -1  1 -1 -1
    
OV(128) =  -1  1 -1 -1  1 -1  1 -1 -1 -1 -1 -1 -1  1 -1  1 -1 -1 -1 -1  1  1 -1  1
    
OV(129) =  -1  1 -1 -1  1 -1  1 -1 -1 -1  1 -1  1  1  1  1 -1 -1 -1 -1 -1  1 -1 -1
    
OV(131) =  -1  1 -1 -1 -1 -1  1 -1 -1 -1  1  1 -1 -1 -1 -1 -1 -1 -1 -1 -1 -1 -1 -1
    
OV(132) =  -1 -1 -1 -1 -1  1  1 -1 -1 -1 -1 -1 -1 -1 -1 -1 -1 -1 -1 -1  1 -1 -1 -1
    
OV(133) =  -1  1 -1 -1  1 -1  1 -1 -1 -1  1 -1  1 -1  1  1 -1 -1 -1 -1 -1  1 -1 -1
    
OV(134) =  -1  1 -1 -1 -1  1  1 -1 -1  1 -1  1 -1  1  1  1 -1 -1 -1 -1  1  1 -1  1
    
OV(136) =  -1  1 -1 -1  1 -1  1 -1 -1 -1  1 -1  1 -1 -1 -1 -1 -1 -1 -1 -1 -1 -1 -1
    
OV(137) =  -1 -1 -1 -1  1 -1  1 -1 -1 -1  1 -1 -1 -1  1  1 -1 -1 -1 -1  1  1 -1  1
    
OV(138) =  -1 -1 -1 -1 -1  1  1 -1 -1 -1 -1 -1 -1  1  1  1 -1 -1 -1 -1  1  1 -1  1
    
OV(139) =  -1  1 -1 -1  1 -1  1 -1 -1 -1 -1 -1 -1  1  1  1 -1 -1 -1 -1  1  1 -1  1
    
OV(141) =  -1 -1 -1 -1  1 -1  1 -1 -1 -1  1 -1  1 -1 -1 -1 -1 -1 -1 -1 -1  1 -1 -1
    
OV(142) =  -1 -1 -1 -1  1 -1  1 -1 -1 -1  1 -1  1 -1 -1 -1 -1 -1 -1 -1 -1 -1  1 -1
    
OV(143) =  -1 -1 -1 -1  1 -1  1 -1 -1 -1  1 -1 -1 -1  1 -1 -1 -1 -1 -1  1  1 -1 -1
    
OV(144) =  -1 -1 -1 -1  1 -1  1 -1 -1 -1  1 -1  1 -1 -1 -1 -1 -1 -1 -1 -1  1  1 -1
    
OV(146) =  -1  1 -1 -1  1 -1  1 -1 -1 -1  1  1  1 -1  1  1 -1 -1 -1 -1 -1 -1 -1 -1
    
OV(147) =  -1  1 -1 -1  1 -1  1 -1 -1 -1  1 -1 -1 -1 -1  1 -1 -1 -1 -1 -1  1 -1 -1
    
OV(148) =  -1 -1 -1 -1  1 -1  1 -1 -1 -1  1 -1 -1 -1  1  1 -1 -1 -1 -1  1  1  1  1
    
OV(149) =  -1  1 -1 -1  1  1  1 -1 -1  1 -1 -1 -1  1  1  1 -1 -1 -1 -1  1  1 -1  1

\medskip{}

(ii) Test Points 29 in number:

Coordinates of Test points:

$PointNo     x_1, x_2,x_3,x_4$ 

 5  5.0  3.6  1.4  0.2 
 
 10 4.9  3.1  1.5  0.1 
 
 15 5.8  4.0  1.2  0.2 
 
 20 5.1  3.8  1.5  0.3 
 
 25 4.8  3.4  1.9  0.2 
 
 30 4.7  3.2  1.6  0.2 
 
 35 4.9  3.1  1.5  0.2 
 
 40 5.1  3.4  1.5  0.2 
 
 45 5.1  3.8  1.9  0.4 
 
 50 5.0  3.3  1.4  0.2 
 
 55 6.5  2.8  4.6  1.5 
 
 60 5.2  2.7  3.9  1.4 
 
 65 5.6  2.9  3.6  1.3 
 
 70 5.6  2.5  3.9  1.1 
 
 75 6.4  2.9  4.3  1.3 
 
 80 5.7  2.6  3.5  1.0 
 
 85 5.4  3.0  4.5  1.5 
 
 90 5.5  2.5  4.0  1.3 
 
 95 5.6  2.7  4.2  1.3 
 
100 5.7  2.8  4.1  1.3 

105 6.5  3.0  5.8  2.2 

110 7.2  3.6  6.1  2.5 

115 5.8  2.8  5.1  2.4 

120 6.0  2.2  5.0  1.5 

125 6.7  3.3  5.7  2.1 

130 7.2  3.0  5.8  1.6 

135 6.1  2.6  5.6  1.4 

140 6.9  3.1  5.4  2.1 

145 6.7  3.0  5.2  2.3 

\medskip{}
OVs of Test Points w.r.t. the 24 planes :

OV(  5) =   1 -1  1  1 -1  1 -1  1  1 -1 -1  1 -1  1 -1 -1  1  1  1  1  1 -1 -1  1

OV( 10) =   1 -1  1  1 -1  1 -1  1  1 -1 -1  1 -1  1 -1  1  1  1  1 -1  1 -1 -1  1

OV( 15) =   1 -1  1  1  1 -1 -1  1  1 -1 -1  1 -1 -1 -1 -1  1  1  1  1  1 -1 -1 -1

OV( 20) =   1 -1  1  1  1  1 -1  1  1 -1 -1  1 -1  1 -1 -1 -1  1  1  1  1 -1 -1  1

OV( 25) =   1 -1  1  1 -1  1 -1  1  1 -1 -1  1 -1  1 -1  1 -1 -1 -1 -1  1 -1 -1  1

OV( 30) =   1 -1  1  1 -1  1 -1  1  1 -1 -1  1 -1  1 -1  1 -1  1  1 -1  1 -1 -1  1

OV( 35) =   1 -1  1  1 -1  1 -1  1  1 -1 -1  1 -1  1 -1  1 -1  1  1 -1  1 -1 -1  1

OV( 40) =   1 -1  1  1 -1  1 -1  1  1 -1 -1  1 -1  1 -1 -1 -1  1  1 -1  1 -1 -1  1

OV( 45) =   1 -1  1  1 -1  1 -1  1  1 -1 -1  1 -1  1 -1 -1 -1 -1 -1 -1  1 -1 -1  1

OV( 50) =   1 -1  1  1 -1  1 -1  1  1 -1 -1  1 -1  1 -1 -1 -1  1  1 -1  1 -1 -1 -1

OV( 55) =  -1  1 -1 -1  1 -1 -1 -1 -1 -1 -1  1 -1 -1 -1  1 -1 -1 -1 -1 -1 -1 -1 -1

OV( 60) =   1  1 -1 -1  1  1 -1  1  1  1 -1  1 -1  1  1  1 -1 -1 -1 -1  1  1 -1  1

OV( 65) =   1  1  1 -1  1 -1 -1  1  1 -1 -1  1 -1  1 -1  1 -1 -1 -1 -1 -1 -1 -1 -1

OV( 70) =   1  1 -1  1 -1 -1 -1  1  1  1 -1  1 -1  1 -1  1 -1 -1 -1 -1  1 -1 -1 -1

OV( 75) =  -1  1 -1 -1 -1 -1 -1  1  1 -1 -1  1 -1 -1 -1  1 -1 -1 -1 -1 -1 -1 -1 -1

OV( 80) =   1  1  1  1  1 -1 -1  1  1 -1 -1  1 -1  1 -1  1 -1 -1 -1 -1 -1 -1 -1 -1

OV( 85) =   1  1 -1 -1  1  1  1  1 -1  1 -1 -1 -1  1  1  1 -1 -1 -1 -1  1  1 -1  1

OV( 90) =   1  1 -1 -1  1 -1 -1  1  1  1  1  1 -1  1  1  1 -1 -1 -1 -1  1  1 -1  1

OV( 95) =   1  1 -1 -1  1  1 -1  1  1  1 -1  1 -1  1  1  1 -1 -1 -1 -1  1  1 -1  1

OV(100) =   1  1 -1 -1  1  1 -1  1  1 -1 -1  1 -1  1 -1  1 -1 -1 -1 -1  1 -1 -1  1

OV(105) =  -1  1 -1 -1  1 -1  1 -1 -1 -1  1 -1 -1  1  1  1 -1 -1 -1 -1  1  1 -1  1

OV(110) =  -1 -1 -1 -1  1 -1  1 -1 -1 -1 -1 -1 -1 -1 -1 -1 -1 -1 -1 -1 -1  1 -1 -1

OV(115) =   1  1 -1 -1  1 -1  1 -1 -1 -1  1 -1  1 -1  1  1 -1 -1 -1 -1 -1  1  1  1

OV(120) =  -1  1 -1 -1 -1 -1  1  1 -1  1  1  1  1  1  1  1 -1 -1 -1 -1  1  1 -1 -1

OV(125) =  -1 -1 -1 -1  1 -1  1 -1 -1 -1 -1 -1 -1 -1 -1 -1 -1 -1 -1 -1  1  1 -1  1

OV(130) =  -1  1 -1 -1 -1  1  1 -1 -1 -1 -1  1 -1  1 -1 -1 -1 -1 -1 -1  1 -1 -1 -1

OV(135) =  -1  1 -1 -1 -1  1  1  1 -1  1 -1  1 -1  1  1  1 -1 -1 -1 -1  1  1 -1  1

OV(140) =  -1 -1 -1 -1  1 -1  1 -1 -1 -1  1 -1 -1 -1 -1 -1 -1 -1 -1 -1 -1 -1 -1 -1

OV(145) =  -1  1 -1 -1  1 -1  1 -1 -1 -1  1 -1  1 -1 -1 -1 -1 -1 -1 -1 -1 -1  1 -1

----------------

\end{document}